%% file: main.tex
\documentclass{article}

\PassOptionsToPackage{numbers, compress}{natbib}

\usepackage[preprint]{neurips_2024}

\usepackage[utf8]{inputenc} 
\usepackage[T1]{fontenc}    
\usepackage{hyperref}       
\usepackage{url}            
\usepackage{booktabs}       
\usepackage{amsfonts}       
\usepackage{nicefrac}       
\usepackage{microtype}      
\usepackage{xcolor}         
\usepackage{multirow}

\usepackage{graphicx}         
\usepackage{amsmath}         
\usepackage{amssymb}         
\AtEndPreamble{
    \usepackage[capitalize]{cleveref}
    \Crefname{section}{Section}{Sections}
    \crefname{section}{Section}{Sections}
    \Crefname{table}{Table}{Tables}
    \crefname{table}{Table}{Tables}
    \Crefname{figure}{Figure}{Figures}
    \crefname{figure}{Figure}{Figures}
}

\title{Enhanced Object Tracking by Self-Supervised Auxiliary Depth Estimation Learning}

\author{%
    Zhenyu Wei \\
    Macau University of Science and Technology\\
    Macao, China \\
    \texttt{2220008267@student.must.edu.mo} \\
    \And
    Yujie He \\
    Macau University of Science and Technology \\
    Macao, China \\
    \texttt{2220006222@student.must.edu.mo} \\
    \And
    Zhanchuan Cai \\
    Macau University of Science and Technology \\
    Macao, China \\
    \texttt{zccai@must.edu.mo} \\
}

\begin{document}
\maketitle
\input{sec/0_abstract}    
\input{sec/1_introduction}
\input{sec/2_related_work}
\input{sec/3_challenges}
\input{sec/4_methods}
\input{sec/5_experiments}
{
    \small
    \bibliographystyle{ieeenat_fullname}
    \bibliography{main}
}
\input{sec/6_appendix}

\end{document}

%% file: sec/0_abstract.tex
\begin{abstract}
RGB-D tracking significantly improves the accuracy of object tracking. However, its dependency on real depth inputs and the complexity involved in multi-modal fusion limit its applicability across various scenarios. The utilization of depth information in RGB-D tracking inspired us to propose a new method, named \textbf{MDE}Track, which trains a tracking network with an additional capability to understand the depth of scenes, through supervised or self-supervised auxiliary \textbf{M}onocular \textbf{D}epth \textbf{E}stimation learning. The outputs of MDETrack's unified feature extractor are fed to the side-by-side tracking head and auxiliary depth estimation head, respectively. The auxiliary module will be discarded in inference, thus keeping the same inference speed. We evaluated our models with various training strategies on multiple datasets, and the results show an improved tracking accuracy even without real depth. Through these findings we highlight the potential of depth estimation in enhancing object tracking performance.
\end{abstract}

%% file: sec/1_introduction.tex
\section{Introduction}
\label{sec:Introduction}

RGB-D tracking utilizes RGB and depth modalities to improve tracking performance. Its core objective is to provide enriched 3D structure cues of scenes to the neural network~\cite{rgbd_review}. While it achieved notable advancements, relying on additional expensive depth acquisition equipments limits its usage scenarios. However, simply substituting estimated depth for the ground truth would require multi-modal fusion modules at the tracking head~\cite{DepthMatters, wang2021depth}, which increases the inference burden, and accuracy of the estimation also influences tracking performance.

Auxiliary learning is to find or design both auxiliary tasks and training methods to enhance the primary task~\cite{liu2022monocon}. When the primary task (e.g., object tracking) struggles to improve its accuracy due to insufficient datasets or other reasons, introducing relevant auxiliary tasks allows enhancing the model's generalization capability~\cite{liebel2018auxiliary, jaderberg2016reinforcement}. It can strengthen the model's generalizability on the primary task by sharing some learnable components~\cite{liu2019self, mordan2018revisiting, liu2022monocon}. This motivates us to seek an \textbf{auxiliary training} method that enhances the performance of object tracking by \textbf{self-supervised depth estimation learning}.

In the study of vision tasks, object tracking and depth estimation are generally recognized as two relatively independent fields, each with unique challenges and solutions. Acquiring training data for depth estimation on a large scale presents significant challenges, and the availability of such data for RGB-D tracking is even more scarce. However, object tracking takes consecutive RGB frames as input, which is suitable for self-supervised depth estimation learning by camera poses prediction and image reconstruction~\cite{monodepth2}. Besides, in RGB-D tracking, depth information is treated as a correlation modality that provides complements for RGB-based tracking~\cite{ViPT,DepthMatters}, thus we argue that depth estimation learning can be considered as an auxiliary task for tracking. 

\begin{figure}
    \centering
    \includegraphics[width=1.0\linewidth]{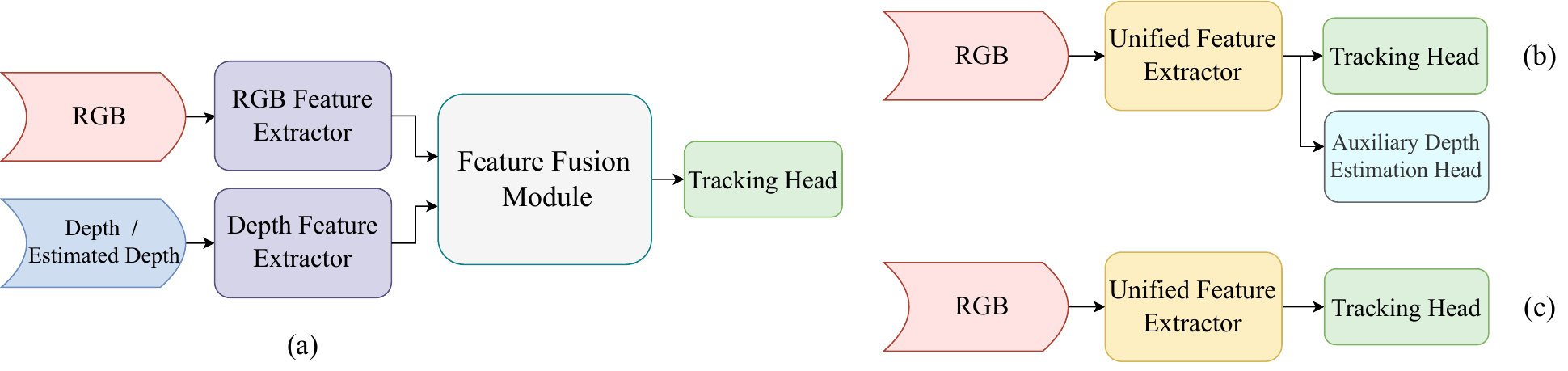}
    \hfill
    \caption{{\bf Comparison between existing multi-modal paradigm and MDETrack.} (a) RGB-D tracking (using real depth) and monocular 3D detection (using estimated depth). (b) MDETrack learns monocular depth estimation through an auxiliary module during training. (c) MDETrack discards the auxiliary module to achieve faster inference.}
    \label{fig:Innovation}
\end{figure}

In this work, we propose MDETrack, a decouplable and efficient end-to-end framework utilizing self-supervised monocular depth estimation learning for enhanced object tracking. We design an auxiliary learning network architecture, which can be added onto an end-to-end object tracking framework. This facilitates the augmentation of the tracker's proficiency in perceiving 3D scenes during training, while allowing for its discarding during inference without compromising tracking efficiency~\cite{liu2022monocon}, as illustrated in \cref{fig:Innovation}. In this way, most RGB videos with consistent camera intrinsics can be trained and inferred without extra information captured by additional devices. We also employ a lightweight visual transformer network to ensure the tracking efficiency. Our contributions are as follows.

\begin{enumerate}
\item We propose an end-to-end framework leveraging depth estimation learning as an auxiliary component for tracking performance enhancement.
\item We unify the preprocessing for self-supervised depth estimation and object tracking, and use a shared feature extraction module to exploit additional spatio-temporal information in both primary task and auxiliary module.
\item We incorporate a self-supervised depth estimation module, which is designed to be discardable post-training. While the system learns from enhanced depth perception during the training phase, it remains efficient at inference time.
\item We show that MDETrack achieves higher accuracy in both supervised and self-supervised learning compared to training for tracking only.
\end{enumerate}


%% file: sec/2_related_work.tex
\section{Related Work}
\label{sec:RelatedWork}

Having witnessed the benefit of depth information to object tracking performance, increasing number of methods start to introduce depth information into the tracker and eventually prove the its importance~\cite{rgbd_review, DepthMatters,ViPT}. We investigate the enhancement of depth information on tracking performance to seek an appropriate method for depth estimation learning and auxiliary learning.

\subsection{Multi-Modal Tracking}
\label{sec:MMT}

Visual object tracking takes a template image and a video frame as input, outputs the position box coordinates of the target referred by the template image~\cite{HiT,stark,ViPT}. And multi-modal tracking methods utilize deep features extracted from RGB images and multi-modal data to provide rich discriminative information for tracking tasks~\cite{DepthMatters}.

Common multi-modal tracking methods bridge upstream feature extraction and downstream tracking modules by a multi-modal feature fusion, utilizing RGB features and depth~\cite{DMTracker, DAL, RGBDAerialTracking, DepthTrack} or other modality~\cite{Fusion_RGBT_Tracking,chiu2021probabilistic} features to fusion modules to extend the tracker and fine-tune it for downstream tracking tasks. Employing these methodologies necessitates the extraction of features for each modality independently, followed by early or late-stage fusion procedures to derive multi-modal features.  More innovative approaches design prompt learning paradigms to finetune pre-trained RGB tracking models, using multi-modal embeddings as prompt tokens to replace extra network branches and improve tracking performance~\cite{ViPT, wang2023review}. In this paper, we draw inspiration from~\cite{ViPT} and design a multi-modal unified feature extractor, which serves to replace both the real depth feature extraction branch and the multi-modal fusion module.

\subsection{Depth Estimation}
\label{sec:DepthEstimation}

Since object tracking takes monocular frame sequences as input, we seek suitable monocular depth estimation methods. Depth estimation predicts depth values by analyzing visual features, motion information and geometric constraints in the images~\cite{eigen2014depth}. Due to the requirement of predicting each pixel in an image, this task is usually classified as a dense prediction task~\cite{DPT, VTM}. Supervised depth estimation needs ground truth depth during training. However, the scarcity of ground truth depth data for training brings a significant challenge. Datasets for RGB-D tracking are even more rare. 

Advanced methods employ self-supervised learning to mitigate the issue of insufficient real depth. Common methods address the problem by using image reconstruction and photometric loss function to train depth estimation models~\cite{Wang_Liang_Xu_Jiao_Yu_2024, LiteMono, monodepth2, monovit, DynaDepth}. Some specific issues in self-supervised depth estimation are addressed in~\cite{chen2023self, he2022ra}. Lite-Mono~\cite{LiteMono} achieves state-of-the-art performance by designing lightweight networks, and adopting a self-supervised image reconstruction module inspired by Monodepth2~\cite{monodepth2}. 

In our work, the problem of image size mismatch will be encountered as our method allows to use separate tracking and depth estimation datasets instead of a single RGB-D tracking dataset. Different image sizes during training and inference would greatly affect the accuracy of depth estimation~\cite{he2022ra}, and in turn, it would affect the accuracy of tracking~\cite{DepthMatters}. Hence we adopt self-supervised depth estimation trained by the RGB frames available from tracking, which eliminates the problem of scarce RGB-D datasets and mismatched image sizes.

\subsection{Depth-Estimation Based Vision Tasks}
\label{sec:DepthEstimationForTracking}

Leveraging estimated depth to enhance the primary task is a common methodology in monocular 3D object detection. After obtaining the estimated depth map, the depth-image based approaches~\cite{liu2024depthaware, wang2021depth, DepthMatters} enhance detection performance by generating depth-aware features through a fusion module that integrates images with depth maps, and pseudo-LiDAR based approaches~\cite{ding2020learning, ma2019accurate, wang2019pseudo} transform depth maps into pseudo-LiDAR point clouds, upon which LiDAR-based detectors are subsequently applied to perform predictions~\cite{mao20233d}. The introduction of camera focal length enables the network to infer depth not only from a single image, but also from the pixel motions and sizes across multiple frames~\cite{park2021pseudo,monodepth2}. These methods all increase the computational burden in both training and inference, and relies on accurate camera intrinsic matrix for depth estimation.

The pioneering work seeks feature fusion modules to replace the pseudo-LiDAR~\cite{mao20233d}. DD3D~\cite{park2021pseudo} finds the training is more stable when performing objective function optimization at multiple scales. MonoCon~\cite{liu2022monocon} and~\cite{chen2023monocular} design auxiliary networks to enable estimation capability for additional modality so that they do not need to explicitly predict depth, the auxiliary architectures are discarded during inference, thus only increasing the burden of training. We pursue a lightweight inference network for efficient tracking performance, therefore in our work we follow MonoCon~\cite{liu2022monocon} to add an auxiliary learning architecture to the tracker for depth estimation learning, which is only applied during training.

%% file: sec/3_challenges.tex
\section{Challenges}
\label{sec:Challenges}

Existing RGB-D tracking methods rely on additional depth acquisition devices to obtain depth information. However, most practical tasks are provided with environments that are lacking sources of depth information, which limits their use in scenarios and more adaptive approaches remain unexplored. As depth estimation methods move towards efficient and lightweight~\cite{LiteMono, hui22rmdepth}, we expect to utilise the features extracted from RGB frames for self-supervised depth estimation learning.

Current RGB tracking methods extract features from input images and feed them into the tracker. During this process, it is unlikely to know whether the network has implicitly estimated depth information from the RGB images to ensure tracking accuracy. Therefore splicing a depth estimation module onto the head of the visual tracker is not the only approach. Alternatively, the network can be trained to extract and utilize multi-modal features, including depth information, necessitating the design of a new architecture. Moreover, this approach is limited by the scarcity of RGB-D datasets. Also, the independent training of depth and tracking could also be affected by factors like scale inconsistency~\cite{multitasklearningoverview}. Thus, design of the network and formulation of the auxiliary learning strategy become pivotal factors in performance of the primary task.

Self-supervised depth estimation depends on the precise estimation of camera motion, which hinges on the effective operation of the camera pose network~\cite{zhou2017unsupervised}. However, the image cropping and resizing common in tracking methods complicate this process. To ensure the effectiveness of the self-supervised depth estimation, we employ image padding to maintain a consistent aspect ratio, without introducing extra geometric distortions. Subsequently, the image edges are filled with solid-colored regions~\cite{stark, HiT}, which still disturb the depth estimation. In addition, the camera tends to move with the objects in tracking tasks, and the network is inclined to predict the relatively stationary pixels in adjacent frames as infinity depth. Such regions and pixels have a negative impact on the self-supervised depth estimation training~\cite{luo2019every, zhou2017unsupervised, monodepth2}. Moreover, the accuracy of self-supervised depth estimation can also be affected by the camera pose. The accuracy of camera pose prediction influences the image reconstruction accuracy. Ideally, the unified training method we design allows the tracking network to learn more generalized knowledge for the feature fusion of RGB and depth for unseen image sequences.


  

%% file: sec/4_methods.tex
\section{Methods}
\label{sec:Methods}

\begin{figure}
    \centering
    \includegraphics[width=1.0\linewidth]{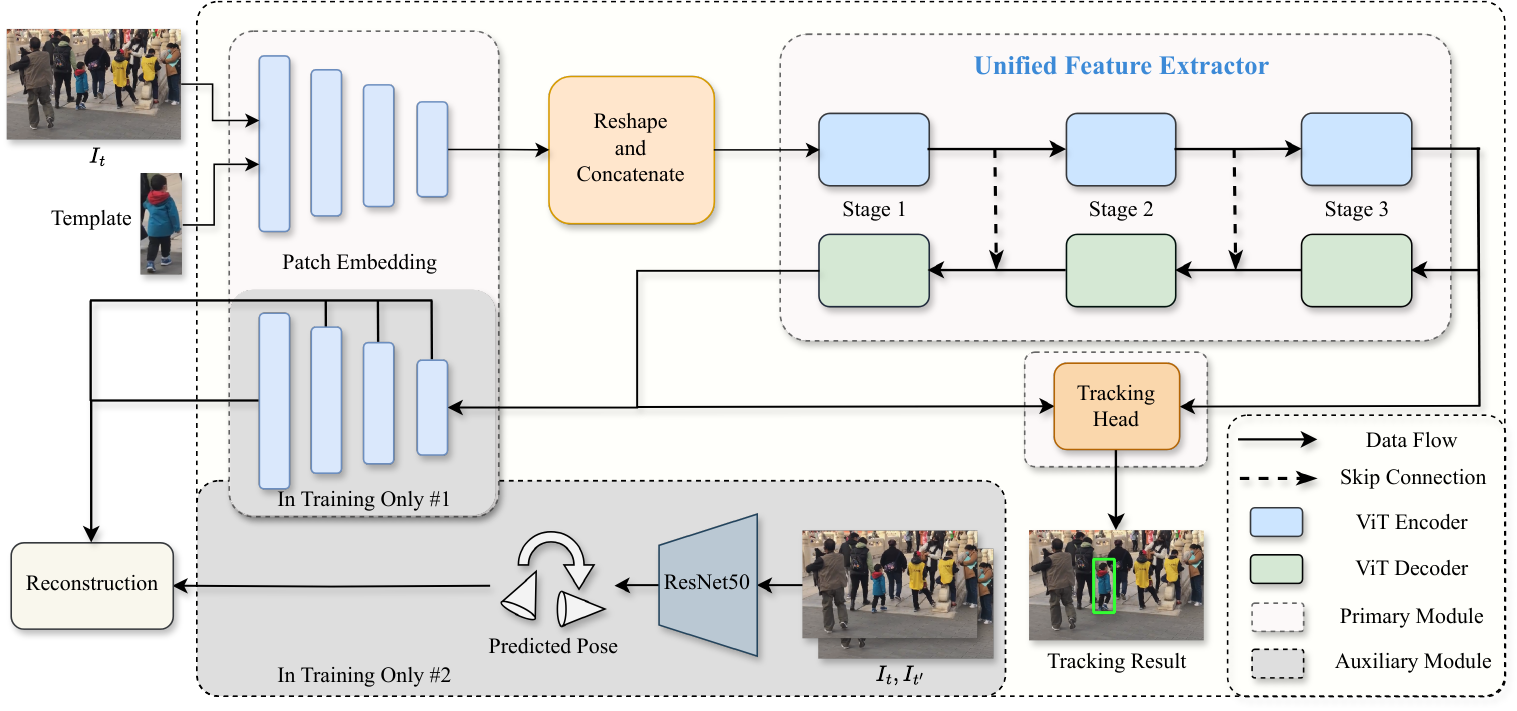}
    \hfill
    \caption{{\bf Overview.} The adjacent frame pair $X=\{I_{t'},I_{t}\}$ generated by data preprocessing is sent for camera pose prediction, and only $I_t$ is sent to the patch embedding module. The dark grey region \#1 is used as an auxiliary learning branch in supervised training, while region \#2 is further enabled for self-supervised auxiliary training.}
    \label{fig:Overview}
\end{figure}

We propose MDETrack, an efficient tracking framework that improves tracking performance by learning monocular depth estimation and extracting multi-modal unified features. Specifically, for training, we unify the data preprocessing for object tracking and depth estimation. And we add a self-supervised monocular depth estimation module on a feature extractor that is suitable for dense prediction tasks. The tracking and depth estimation heads share the extractor for multi-modal unified feature extraction. The depth estimation head will be discarded in inference to ensure the models have consistent parameters. 

In \cref{sec:Preprocessing}, we describe how we perform data preprocessing to accommodate the primary and auxiliary tasks and mitigate the effects of stationary pixels from tracking on self-supervised depth estimation. We give an overview of the proposed MDETrack components in \cref{sec:MDE_Track}. In \cref{sec:Supervised} and \cref{sec:Self_Supervised}, we discuss our supervised and self-supervised auxiliary learning methods.

\subsection{Preprocessing}
\label{sec:Preprocessing}

The object tracking and depth estimation tasks differ significantly when it comes to image preprocessing. Since tracking is our primary task, we bias the preprocessing towards tracking, and prevent performance degradation of self-supervised depth estimation due to this bias.

\paragraph{Camera Pose} The training of self-supervised depth estimation is based on accurate camera pose prediction, which seeks to predict the 6-DoF relative pose of the camera by a pair of color frames when capturing a static scene~\cite{monodepth2}. The tracking datasets provide a more stable motion for dynamic object tracking, but also provide an uncertainty in the camera motion between frame pairs. Therefore, for each image $I_t$ in \cref{fig:Overview} as a base frame, we randomly select one frame from both previous and next frames, respectively, according to temporal order. The selection process is formulated as
\begin{equation}
    P_{d}=\{i'\mid \min(t,t+dR)< i'<\max(t,t+dR),i'\in P\},
\end{equation}
\begin{equation}
    t'_d=\begin{cases} 
        \text{Random}(P_{d}),&\text{if }|P_{d}|\neq0,
         \\ 
         t,&\text{otherwise,} 
        \end{cases}
    \label{eq:RandomSelect}
\end{equation}
where $d\in\{-1,1\}$, and $P_{d}$ is the set of selectable frames in the direction $d$. $t'_d$ must be within the valid frames set $P$, which consists of in-plane labeled frames, and we set the selection range $R=5$. If there is no available frame in set $P_{d}$ for any $d$, the current frame is used instead, i.e. $t'_d=t$. We use Resnet50~\cite{resnet} as the camera pose prediction network following the previous works~\cite{monodepth2,LiteMono,godard2017unsupervised}. Each of the two input image pairs is $X=\{I_{t'},I_{t}\}$ where $t'\in\{t'_{-1},t'_{1}\}$, indicating previous or next two frames arranged in temporal order. The input image augmentation is aligned with~\cite{HiT}, including image normalization, transform to grayscale with 5\% probability, jitter with 20\% probability, and image flip with 50\% probability.

\begin{figure}
    \centering
    \includegraphics[width=1.0\linewidth]{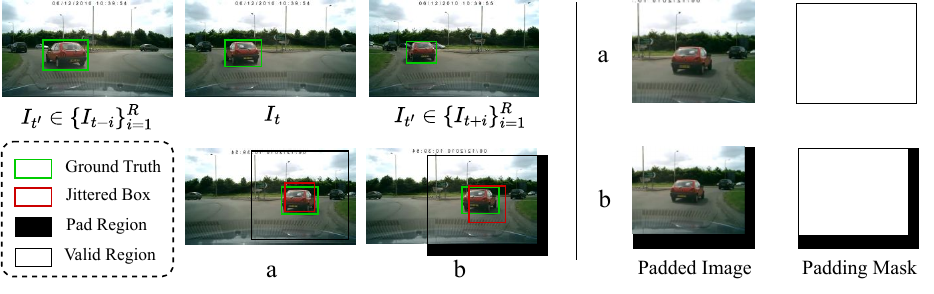}
    \caption{{\bf Left:} The first row demonstrates the way to compose a sequence of frames $X=\{I_{t'},I_{t}\}$ by random sampling, with $R$ as the sampling range. The second row shows the padding mask generation process. {\bf Right:} Relationship between the padded image and padding mask.}
    \label{fig:PreProcessing}
\end{figure}

\paragraph{Search Region and Template Image}\label{SearchRegionandTemplateImage} The template image identifies the object to be tracked, the search region is a cropped region for each frame in the same video sequence during data preprocessing. As shown in \cref{fig:PreProcessing}, after jittering the centre and size of the target box, the search and template image are generated with the jittered box as the centre. Images $a$ and $b$ illustrate the image augmentation including random flip, and the respective padding masks generated after different jitters of the target box, indicated by the black regions. In image preprocessing, aspect ratio of the image is not changed to ensure effectiveness of the self-supervised depth estimation, as well as to avoid introducing extra geometric distortions. The border region of the image is first padded to the crop size with a crop factor, then a square crop centred on the target box is extracted and then resized to the target size. The size of all search images of $H\times W$ and template images of $h\times w$ are cropped and padded to $256\times 256$ and $128\times 128$, respectively. This can also affect the depth estimation learning due to the black region. Hence, we utilize a padding mask to mark the padded region as invalid, contrast to the valid original search region. This area will be filtered out in the supervised depth estimation loss and the self-supervised image reconstruction loss.

\paragraph{Padding Mask} \cref{fig:PreProcessing} illustrates the process of image cropping and padding, where cropped image smaller than the target size will be padded and a padding mask $m_p$ will be generated. In $m_p$, we assign 0 to the unpadded valid region after image padding and 1 to the padded region.

\subsection{MDETrack}
\label{sec:MDE_Track}

\cref{fig:Overview} shows the proposed architecture of MDETrack. It is comprised of three components: a unified feature extractor, a tracking head and a depth estimation head. The dark grey regions \#1 and \#2 are activated as auxiliary regions only for training. Region \#1 is used only for supervised auxiliary learning, while region \#2 is further enabled for self-supervised auxiliary training. The auxiliary region will be discarded in inference. The depth estimation head is located in region \#1 and \#2, and is designed for the auxiliary learning. Intuitively, the auxiliary modules provide additional 3D structure information to the network, which is beneficial for tracking tasks. Experiments in \cref{sec:Experiments} demonstrate its capability to enhance the tracking performance. Patch embedding and tracking head are detailed in \cref{sec:ImplementationDetails}.

\paragraph{Unified Feature Extractor} To retain the high inference speed enabled by auxiliary learning, we draw inspiration from~\cite{HiT,LiteMono} to design the multi-modal unified feature extractor as a lightweight hierarchical vision transformer~\cite{LeViT}, and introduce skip connections to keep details. The downsampling encoder and upsampling decoder featured in \cite{LeViT} make it a suitable backbone for dense prediction tasks~\cite{LiteMono}. The adjacent frame pair $X$ is generated after data preprocessing, and only the middle frame $I_t$ in $X$ is fed to the unified feature extractor. We train the extractor to predict multi-modal unified features by auxiliary learning. As shown in \cref{fig:Overview}, $I_t$ and template image entering the network are reshaped and concatenated after patch embedding, following which they are fed to the extractor to get the image features $\phi$. Skip connections are added so the feature information at different resolution scales can be passed directly from encoders at different stages to the corresponding decoder, thus preserving the feature details.

\paragraph{Depth Estimation Head}\label{DepthEstimationHead} We simply apply an architecture that is similar to an inversed patch embedding as the depth estimation head $\mathrm{H}_{\text{d}}$. $\mathrm{H}_{\text{d}}$ is added onto the end of the feature extractor parallel with the tracking prediction head. 

\subsection{Supervised Auxiliary Learning}
\label{sec:Supervised}

We follow~\cite{HiT} in designing the unified feature extraction module. Specifically, the feature extractor $\mathrm{E}: I^{H \times W \times 3} \oplus I^{h \times w \times 3} \to \phi$ takes dual-image position encoded frames, denoted as $I_\varepsilon$, as the input and extracts image features $\phi$. Tracking head $\mathrm{H}_{\text{tk}}:\phi \to B^{4}$ predicts boxes $B$ from $\phi$, and depth estimation head $\mathrm{H}_{\text{d}}:\phi \to I^{H \times W \times 1}$ estimates depth image. The overall workflow is 
\begin{equation}
    \mathrm{F}(I_t) = (\mathrm{H}_{\text{tk}}\circ \mathrm{E})(I_\varepsilon)+(\mathrm{H}_{\text{d}}\circ \mathrm{E})(I_\varepsilon).
    \label{eq:SupervisedWorkflow}
\end{equation}
We use padding mask to obviate the effect of image padding on depth estimation. When computing the supervised depth estimation loss $L_d$, we focus only on the unpadded valid region where $m_p$ is 0,
\begin{equation}
    L_{sd}(\hat{D}(I_t;\theta),D,m_p)=\Vert(1-m_p)\cdot (\hat D-D)\Vert_1,
    \label{eq:SupervisedMask}
\end{equation}
where $\hat{D}$ and $D$ are the predicted and real depth, respectively, and $\theta$ denotes the learnable parameters in the overall supervised auxiliary learning network. The supervised depth estimation loss $L_{sd}$ is the L1 loss between real and predicted depth images. The total loss for supervised auxiliary training is 
\begin{equation}
    L=(1-\alpha)L_{tk}(\hat{B}(I_\varepsilon;\theta),B)+\alpha L_{sd}.
    \label{eq:SupervisedLoss}
\end{equation}
The loss for tracking is formulated as $L_{tk}=\lambda_{G}L_{G}(\hat{B},B)+\lambda_{l}L_{l}(\hat{B},B)$, where $L_{G}$ and $L_{l}$ are GIoU loss~\cite{rezatofighi2019generalized} and L1 loss, $\hat{B}$ and $B$ are tracking groundtruth and prediction boxes, $\lambda_{G}$, $\lambda_{l}$ and $\alpha$ are hyperparameters, respectively~\cite{HiT}. In experiments, we set $\lambda_{G}=2$ , $\lambda_{l}=5$ and $\alpha=0.4$. We performed supervised learning to evaluate the impact that depth estimation learning provides for the tracking performance. Experimental results are shown in \cref{sec:ExperimentsSupervised}. 

\subsection{Self-Supervised Auxiliary Learning}
\label{sec:Self_Supervised}

\paragraph{Camera Pose Prediction}\label{CameraPosePrediction} As the tracking datasets employ different camera specifications, we use a pseudo camera intrinsic matrix $K$ with equal aspect ratio to predict the camera pose on square frames. The square frames come from image cropping and padding in \cref{SearchRegionandTemplateImage}, which preserves the original aspect ratio of the image and adjusts the frame to square. The ResNet50~\cite{resnet} pretrained on ImageNet~\cite{imagenet} is adapted to receive an adjacent image pair $X$, to predict the camera 6-DoF relative pose $T_{t\to t'}$. The camera pose prediction is trained simultaneously as an auxiliary module of tracking task, by minimizing the image reconstruction loss $L_p$~\cite{zhou2017unsupervised,monodepth2}.

\paragraph{Image Reconstruction}\label{ImageReconstruction} We keep the predicted depth images $\hat{D}$ at different scales ($2^{-\omega}, \omega \in \{0,1,2,3\}$) obtained by each upsampling layer in the depth estimation head, and input them into the image reconstruction module for multi-scale depth estimation~\cite{monodepth2,scharstein2002taxonomy}, to stabilize the performance of self-supervised depth estimation~\cite{park2021pseudo}. We upsample the multi-scale $\hat D$ to the same size as $I_t$, and compute the photometric reprojection image with $T_{t\to t'}$~\cite{jaderberg2015spatial}, which involves $K$. Inspired by \cite{monodepth2}, the reconstructed image is computed by the grid sampling of $I_{t'}$ and the reprojected image,

\begin{equation}
    I_{t'\to t}=F_{sample}(I_{t'},F_{proj}(\hat{D},T_{t\to t'}(X,K;\theta))),
    \label{eq:CameraPose}
\end{equation} 

where $I_{t'\to t}$ denotes the reconstructed $I_t$ from the camera pose and the depth map of $I_{t'}$. $F_{sample},F_{proj}$ stand for the sampling and reprojection operator, respectively, and $\theta$ is the learnable parameters in the self-supervised auxiliary learning network. We follow \cite{godard2017unsupervised} in using SSIM~\cite{wang2004image} and L1 loss to calculate the image reconstruction loss, and the weight $\beta$ is set to 0.85. The padding region is excluded from the calculation by applying the padding mask

\begin{equation}
    L_p=(1-m_p)\sum_{t'}\left(\beta\frac{1-\text{SSIM}(I_{t'\to t},I_t)}{2}+(1-\beta)\Vert I_{t'\to t}-I_t\Vert_1 \right).
    \label{eq:ReprojectionLoss}
\end{equation} 

\paragraph{Optimization}\label{Optimization} We employ a stationary mask~\cite{monodepth2} denoted as $m_s$ to selectively filter out pixels that exhibit minimal relative motion with respect to the camera, thus to mitigate the impact of simultaneous movement between tracking target and camera on self-supervised depth estimation. Furthermore, we incorporate the concept of minimum reprojection loss~\cite{monodepth2} to tackle out-of-view pixels and occluded targets within the source image,

\begin{equation}
    m_s=\min_{t'\in \{t'_{-1},t'_1\}}L_p(I_t,I_{t'\to t})< \min_{t'\in \{t'_{-1},t'_1\}}L_p(I_t,I_{t'}).
    \label{eq:mu}
\end{equation} 

Furthermore, we follow~\cite{monodepth2} in refining the edges of the generated depth maps by incorporating a smoothing loss $L_{s} = |\partial_xd^*_t|e^{-|\partial_xI_t|} +|\partial_yd^*_t|e^{-|\partial_yI_t|}$. Therefore, the overall self-supervised depth estimation auxiliary learning loss is defined as
\begin{equation}
    L=(1-\alpha)L_{tk}(\hat{B}(I_\varepsilon;\theta),B)+\alpha L_{ssd},
    \label{eq:SelfSupervisedLoss}
\end{equation} 
where $\theta$ is the learnable parameters and $\alpha$ is set to $0.4$. $L_{ssd}=m_s L_p + \lambda L_s$ is the loss of self-supervised depth estimation, where we set $\lambda$ to $1\times 10^{-3}$. Experimental results are shown in \cref{sec:ExperimentsSelfSupervised}.


%% file: sec/5_experiments.tex
\section{Experiments}
\label{sec:Experiments}

In view of the variability of camera specifications in the tracking dataset, we employ an equal aspect ratio pseudo-camera intrinsic matrix as $K$ for the input images filled with squares, which are used for self-supervised depth estimation learning. The model is trained for camera pose prediction under the assumption that the input images has no significant distortion. As shown in \cref{tab:Strategy} we formulate four training strategies. In our study, strategy \#1 and \#2 serve as baseline configurations, focusing exclusively on tracking without auxiliary learning. To assess the impact of self-supervised auxiliary learning, we compare strategy \#1 and \#4. Meanwhile, the performance of supervised auxiliary learning is evaluated through a comparison between strategy \#2 and \#3. Code is available at \url{https://anonymous.4open.science/r/MDETrack}.

\begin{table}[htbp!]
    \centering
    \caption{Training Strategies.}
    \begin{tabular}{@{}lll@{}}
    \toprule
    Method                           & \# Strategy & Training Datasets       \\ \midrule
    \multirow{2}{*}{Tracking}        & 1  & LaSOT + GOT-10K + DepthTrack \\
                                     & 2  & DepthTrack              \\
    Supervised                       & 3  & DepthTrack              \\
    Self-Supervised                  & 4  & LaSOT + GOT-10K + DepthTrack \\ \bottomrule
    \end{tabular}
    \label{tab:Strategy}
\end{table}

\begin{figure}
    \centering
    \includegraphics[width=1.0\linewidth]{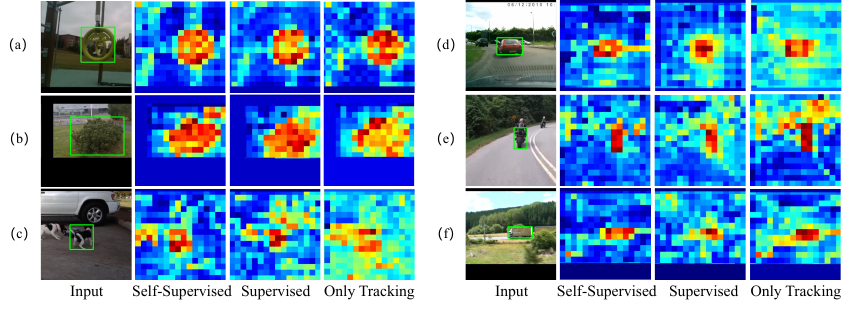}
    \caption{Visualization of the attention maps in tracking head. Networks with auxiliary depth estimation possess an improved capability to discern the 3D structure of the scene, thereby enabling a more concentrated focus on the expected target.}
    \label{fig:Map}
\end{figure}

\subsection{Supervised Auxiliary Learning}
\label{sec:ExperimentsSupervised}

For strategies 2 and 3, we choose the training set of DepthTrack~\cite{DepthTrack} as our dataset for the supervised auxiliary learning and set strategy 2 as the baseline. In this section, we do not explore using more training data. The input to the network consists of 3 parts: template and search image pairs obtained from the RGB images in the dataset, and real depth images sourced from the depth data. Depth images are used to compare with the predicted depth to calculate the depth estimation losses. We use~\cite{LeViT}, which is pretrained on ImageNet~\cite{imagenet}, as the initialized unified feature extractor. Other training details follow the setup of \cite{HiT}, with AdamW~\cite{AdamW} as the optimizer, the weight decay of $1\times 10^{-4}$, and the learning rate set to $5\times 10^{-4}$. We train the model for 300 epochs on one Nvidia RTX 3080 Ti GPU, with the learning rate reduced by 10 times after 240 epochs. Each epoch contains 5000 samples and the batch size is set to 16.

\subsection{Self-supervised Auxiliary Learning} 
\label{sec:ExperimentsSelfSupervised}

For the static pseudo camera intrinsic outlined in \cref{CameraPosePrediction}, the inconsistencies in image distortions across the training datasets result in imprecise predictions of camera pose, which subsequently undermine the effectiveness of self-supervised learning for depth estimation. Consequently, it is imperative that the images used for training exhibit minimal significant lens distortion. For instance, TrackingNet~\cite{Muller_2018_ECCV} contains a wide range of videos captured by cameras with a variety of field-of-view (FOV) values. Therefore we selectively filtered the common tracking datasets and use only LaSOT~\cite{fan2019lasot}, GOT-10K~\cite{huang2019got} and DepthTrack~\cite{DepthTrack} for training. For strategies 1 and 4, each epoch contains 15,000 samples and the batch size is set to 16. We set strategy 1 as the baseline. Other training settings are identical to \cref{sec:ExperimentsSupervised}.

\subsection{Results and Analysis} 
\label{sec:ResultsandAnalysis}

We test the performance of the models for each training strategy. We evaluate the performance of the models on the test sets of LaSOT, GOT-10K, DepthTrack, and VOT-RGBD2022~\cite{kristan2022tenth}, and also measure the number of parameters of the models as well as FPS at testing time. All tests are performed on a single Nvidia RTX 3080 Ti GPU and Intel(R) Xeon(R) Silver 4214R CPU @ 2.40GHz. \cref{tab:Results} shows the results.

\paragraph{LaSOT} LaSOT~\cite{fan2019lasot} is a large-scale RGB tracking dataset that features 85 balanced categories, predominantly comprised of data captured by cameras in motion. In our evaluation, as detailed in \cref{tab:Results}, the self-supervised MDETrack demonstrates enhanced performance over the established baseline of 1.11\%, 1.19\%, and 1.23\% on AUC, P, and P$_{\text{Norm}}$, respectively. The supervised method also outperforms the baseline by 0.75\%, 1.24\%, and 1.41\% in the corresponding metrics.

\paragraph{GOT-10K} GOT-10K~\cite{huang2019got} is a large-scale RGB tracking dataset with most of its data captured through cameras in motion. The self-supervised MDETrack surpasses the baseline with a marginal improvement of 0.3\% in the Success Rate at a threshold of 0.75 (SR$_{0.75}$). However, it slightly underperforms in terms of Average Overlap (AO) and Success Rate at a threshold of 0.5 (SR$_{0.5}$), with decreases of 0.5\% and 0.7\%, respectively, compared to the baseline. The supervised method exhibits a reduction in performance across all metrics.

\paragraph{DepthTrack} DepthTrack~\cite{DepthTrack} adds cases for RGB-D tracking tasks such as dark scenes, similarity color and texture between target and background, deformed and similar objects. We observe that most of the data is captured by fixed cameras. The self-supervised MDETrack is superior than the baseline by 3.6\%, 3.9\% and 3.8\% on Precision, Recall and F-score, respectively. The supervised method also outperforms the baseline by 1.3\% in Precision, 1.5\% in Recall, and 1.4\% in F-score.

\paragraph{VOT-RGBD2022} VOT-RGBD2022~\cite{kristan2022tenth} presents a collection of short-term tracking scenarios, and most of its data are captured by fixed cameras. MDETrack achieves enhancements by 3.4\%, 2.3\% and 3.0\% on EAO, A and R, respectively, over the baseline. There are slight decreases in the supervised method.

To gain better insights into the impact of auxiliary depth estimation learning on the network's performance, we visualize the attention maps within the tracking head, as illustrated in \cref{fig:Map}. We find that with the support of auxiliary depth estimation learning, the networks achieve a better 3D structure comprehension of scenes. This capability allows the network to adeptly distinguish between objects that resemble the target but are situated at varying depths. Consequently, it enhances the network's focus on accurately identifying the correct target.

\begin{table}[t]
    \centering
    \large
    \caption{Performance on the training strategies. \#1 and \#2 are baselines for training without auxiliary learning. \#1 and \#4 show the impact of self-supervised auxiliary learning, while \#2 and \#3 show impact of supervised auxiliary learning. The best results are highlighted in {\color[HTML]{FE0000} red}.}
    \resizebox{\columnwidth}{!}{%
    \begin{tabular}{@{}cccccccccccccccc@{}}
    \toprule
    \multirow{2}{*}{\#}    & \multicolumn{3}{c}{LaSOT} &  & \multicolumn{3}{c}{GOT-10K}      &  & \multicolumn{3}{c}{DepthTrack} &  & \multicolumn{3}{c}{VOT-RGBD2022} \\ \cmidrule(lr){2-4} \cmidrule(lr){6-8} \cmidrule(lr){10-12} \cmidrule(l){14-16} 
                           & AUC     & P      & NP     &  & AO   & SR$_{0.5}$ & SR$_{0.75}$ &  & Pr       & Re      & F-score   &  & EAO       & A         & R        \\[0.05cm]\midrule
    \multicolumn{1}{c|}{1} & 57.94   & 57.37  & 64.90  &  & {\color[HTML]{FE0000} 59.6} & {\color[HTML]{FE0000} 68.0} & 48.7 &  & 0.435    & 0.452   & 0.443     &  & 0.481     & 0.722     & 0.625    \\[0.1cm] 
    \multicolumn{1}{c|}{2} & 31.88   & 24.21  & 32.76  &  & 34.5 & 37.1       & 17.3        &  & 0.339    & 0.352   & 0.345     &  & 0.298     & 0.643     & 0.445    \\[0.15cm]
    \multicolumn{1}{c|}{3} & 32.63   & 25.45  & 34.17  &  & 32.1 & 34.6       & 14.9        &  & 0.352    & 0.367   & 0.359     &  & 0.292     & 0.624     & 0.438    \\[0.15cm]
    \multicolumn{1}{c|}{4} & {\color[HTML]{FE0000} 59.05}   & {\color[HTML]{FE0000} 58.56}  & {\color[HTML]{FE0000} 66.13}  &  & 59.1 & 67.3       & {\color[HTML]{FE0000} 49.0}        &  & {\color[HTML]{FE0000} 0.471}    & {\color[HTML]{FE0000} 0.491}   & {\color[HTML]{FE0000} 0.481}     &  & {\color[HTML]{FE0000} 0.515}     & {\color[HTML]{FE0000} 0.745}     & {\color[HTML]{FE0000} 0.655}      \\ [0.05cm]\bottomrule
    \end{tabular}%
    }
    \label{tab:Results}
    \end{table}

\section{Conclusion}
\label{sec:Conclusion}

This paper introduces MDETrack, a novel framework that utilizes supervised or self-supervised learning for depth estimation as an auxiliary task. This integration enables the network with depth perception capabilities. The supervised method expose lower generalizability on small datasets. However, our proposed self-supervised method, MDETrack, extends the auxiliary depth estimation learning to any tracking dataset with consistent camera intrinsics, and achieves an improvement in accuracy. Remarkably, MDETrack achieves these advancements without compromising on inference speed. The experimental results clearly demonstrate that incorporating the auxiliary depth estimation training improves tracking performance across the majority of test datasets. This demonstrates that the integration of depth estimation capabilities exerts benefits on the efficacy of tracking networks.

\paragraph{Limitations}\label{sec:Limitations} Despite not requiring real depth data during both training and testing phases, the performance of our method still relies on the camera's movement and a known camera intrinsic. Given the variability in camera specifications across the tracking datasets, the quality of the estimated depth may not be good enough in all cases. Additionally, sequences within the training data where the camera remains static also negatively influence the performance of self-supervised depth estimation.

%% file: sec/6_appendix.tex
\appendix

\section{Appendix / supplemental material}

\subsection{Implementation Details}
\label{sec:ImplementationDetails}

\paragraph{Patch Embedding}The patch embedding designed by~\cite{LeViT} with 4 layers of 2 strides $3\times 3$ convolutions reduces the image resolution by a downsampling factor of 16, thus reduces the computational complexity of the transformer's lower layers while improving accuracy. 

\paragraph{Tracking Head}We follow~\cite{HiT} to compute the attention map from the output of stage 3 of the encoder along with the output of the decoder. We show that the depth estimation auxiliary module helps the network to focus on the tracked object in \cref{fig:Map} by observing the change in the attention map. The attention map and $\phi$ are sent to the corner head to compute the predict box.

\subsection{Model Metrics}
\label{sec:Metrics}

To confirm that auxiliary learning introduces no additional computational burden, we conducted comprehensive tests across all models. The findings, as detailed in \cref{tab:Speed}, reveal that both the number of parameters in the backbone (\#1) and the tracking head (\#2) remain consistent across models. And the running speeds, measured in FPS, are nearly identical, affirming the efficiency of our approach. Running speeds are tested on one Nvidia RTX 3080 Ti GPU and Intel(R) Xeon(R) Silver 4214R CPU @ 2.40GHz. The results show the models' performance meets the real-time threshold of 20 FPS, as established by the VOT real-time setting~\cite{kristan2022tenth}.

\begin{table}[h]
    \centering
    \caption{Parameter count and speed test results. The models of all training strategies have the same number of parameters and similar running speeds.}
    \begin{tabular}{@{}ccccc@{}}
    \toprule
    \#1 params(M)  & \#2 params(M)  & GPU & CPU \\\midrule
    37.526         & 4.614          & 108 & 19\\ \bottomrule
    \end{tabular}
    \label{tab:Speed}
\end{table}

\subsection{Licenses for Existing Assets}
\label{sec:Licenses}

The code, data, models used in this work are shown below.

\paragraph{Code} Our code is implemented based on HiT\footnote{\url{https://github.com/kangben258/HiT}} (MIT License), ViPT\footnote{\url{https://github.com/jiawen-zhu/ViPT}} (MIT License), and Monodepth2\footnote{\url{https://github.com/nianticlabs/monodepth2}} (Monodepth v2 License).

\paragraph{Data} The datasets used in this work is from LaSOT\footnote{\url{http://vision.cs.stonybrook.edu/~lasot/}} (Apache-2.0 license), GOT-10K\footnote{\url{http://got-10k.aitestunion.com/}} (CC BY-NC-SA 4.0), DepthTrack\footnote{\url{https://github.com/xiaozai/DeT}} (GPL-3.0 license),  VOT-RGBD2022\footnote{\url{https://www.votchallenge.net/vot2022/}} (GPL-3.0 license), and TrackingNet\footnote{\url{https://tracking-net.org/}} (Apache License 2.0).

\paragraph{Models} The models used in this work are from LeViT\footnote{\url{https://github.com/facebookresearch/LeViT}} (Apache-2.0 license) and ResNet\footnote{\url{https://pytorch.org/vision/main/models/resnet.html}} (BSD-3-Clause license), pretrained on ImageNet~\cite{imagenet} (BSD-3-Clause license).

\subsection{Broader Impacts}
\label{sec:BroaderImpacts}

The self-supervised auxiliary learning methodology proposed in MDETrack addresses the challenge of data scarcity and generalizability, which are critical hurdles in machine learning and computer vision. By demonstrating that self-supervised learning can effectively leverage existing datasets without the need for extensive labeled data, our research encourages the adoption of more efficient and accessible machine learning methodologies. This could accelerate the development of intelligent systems in resource-constrained environments and promote broader adoption of AI technologies.

Given the subject of enhancing object tracking, potential areas of negative concern include privacy infringements through surveillance or biases in tracking accuracy due to dataset imbalances.